\documentclass[final]{cvpr}

\usepackage{times}
\usepackage{epsfig}
\usepackage{graphicx}
\usepackage{amsmath}
\usepackage{amssymb}
\usepackage{setspace}
\usepackage{pythonhighlight}
\usepackage{etoolbox}

\usepackage{inconsolata}

\usepackage{multirow}
\usepackage{rotating}
\usepackage{booktabs}

\usepackage[olditem,oldenum]{paralist}

\usepackage{mysymbols}

\usepackage{url}
\usepackage{xspace}
\usepackage{comment}
\usepackage{color}
\usepackage{xcolor}
\usepackage{framed}
\usepackage{fancybox}
\usepackage{caption}
\usepackage{paralist}

\usepackage{footnote}
\makesavenoteenv{tabular}

\definecolor{citecolor}{HTML}{2779af}
\definecolor{linkcolor}{HTML}{c0392b}
\usepackage[pagebackref=false,breaklinks=true,colorlinks,bookmarks=false,citecolor=citecolor,linkcolor=linkcolor]{hyperref}

\newcommand{\citet}[1]{\cite{#1}}
\newcommand{\citep}[1]{\cite{#1}}

\newcommand{\myapprox}{{\raise.17ex\hbox{$\scriptstyle\sim$}}}
\newcommand{\xhdr}[1]{\vspace{0pt}\noindent\textbf{#1}\xspace}

\newcommand{\reffig}[1]{Fig.~\ref{#1}}
\newcommand{\refsec}[1]{Sec.~\ref{#1}}
\newcommand{\reftab}[1]{Tab.~\ref{#1}}
\newcommand{\refapx}[1]{Apx.~\ref{#1}}
\makeatletter
\newcommand\footnoteref[1]{\protected@xdef\@thefnmark{\ref{#1}}\@footnotemark}
\makeatother

\newcommand{\rgbd}{\texttt{RGB-D}\xspace}
\newcommand{\rgb}{\texttt{RGB}\xspace}
\newcommand{\depth}{\texttt{Depth}\xspace}

\newcommand{\pointnav}{\texttt{PointGoalNav}\xspace}
\newcommand{\pointnavfull}{PointGoal Navigation\xspace}
\newcommand{\compassgps}{GPS+Compass\xspace}
\newcommand{\gpscompass}{\compassgps}

\newcommand{\savva}{Savva \etal~\cite{habitat19iccv}\xspace}
\newcommand{\ddppo}{Wijmans \etal~\cite{ddppo}\xspace}
\newcommand{\both}{\savva; \ddppo}

\belowdisplayskip \abovedisplayskip

\newcommand{\csection}[1]{
    \vspace{-0.08in}
    \section{#1}
    \vspace{-0.07in}
}

\newcommand{\csubsection}[1]{
    \vspace{-0.09in}
    \subsection{#1}
    \vspace{-0.08in}
}

\def\cvprPaperID{4888} %
\def\confYear{CVPR 2021}

\newtoggle{supmain}

\begin{document}

\title{How to Train PointGoal Navigation Agents \\on a (Sample and Compute) Budget}

\author{Erik Wijmans$^{\texttt{1},\texttt{2}}$, Irfan Essa$^{\texttt{1},\texttt{3}}$, and Dhruv Batra$^{\texttt{2},\texttt{1}}$ \\
${^\texttt{1}}$Georgia Institute of Technology,
${^\texttt{2}}$Facebook AI Research, ${^\texttt{3}}$Google Research Atlanta\\
\texttt{\{etw, irfan, dbatra\}@gatech.edu}}

\maketitle

\begin{abstract}
  \pointnavfull has seen significant recent interest and progress, spurred on by the Habitat platform and associated challenge~\citep{habitat19iccv}.
In this paper, we study \pointnavfull under both a \emph{sample} budget (75 million frames) and a \emph{compute} budget (1 GPU for 1 day).  
We conduct an extensive set of experiments, cumulatively totaling over 50,000 GPU-hours, that let us identify and discuss a number of
ostensibly minor but significant design choices -- the advantage estimation procedure (a key component in training), visual encoder architecture, and a seemingly minor hyper-parameter change.
Overall, these design choices to lead considerable and consistent improvements over the baselines present in \savva. 
Under a sample budget, performance for \rgbd agents improves 8 SPL on Gibson (14\% relative improvement) and \emph{20} SPL on Matterport3D (38\% relative improvement). Under a compute budget, performance for \rgbd agents improves by 19 SPL on Gibson (32\% relative improvement) and \emph{35} SPL on Matterport3D (220\% relative improvement). We hope our findings and recommendations will make serve to make the community's experiments more efficient.
\end{abstract}

\csection{Introduction}

Galvanized by fast simulation platforms~\citep{habitat19iccv,xia2020interactive,savva2017minos}, large rich 3D datasets~\citep{replica,mp3d,xia2018gibson}, and the success of deep reinforcement learning~\citep{silver2017mastering,tian2019elf,OpenAI_dota}, training virtual robots (embodied agents) in simulation has garnered considerable interest in recent years.  Works have developed a rich set of tasks, ranging from \pointnavfull \citep{anderson2018evaluation}, to grounded instruction following~\citep{anderson2018vision,shridhar2020alfred} and question answering~\citep{embodiedqa,gordon2018iqa,wijmans2019embodied}.

\begin{figure}
    \centering
    \includegraphics[width=0.95\linewidth]{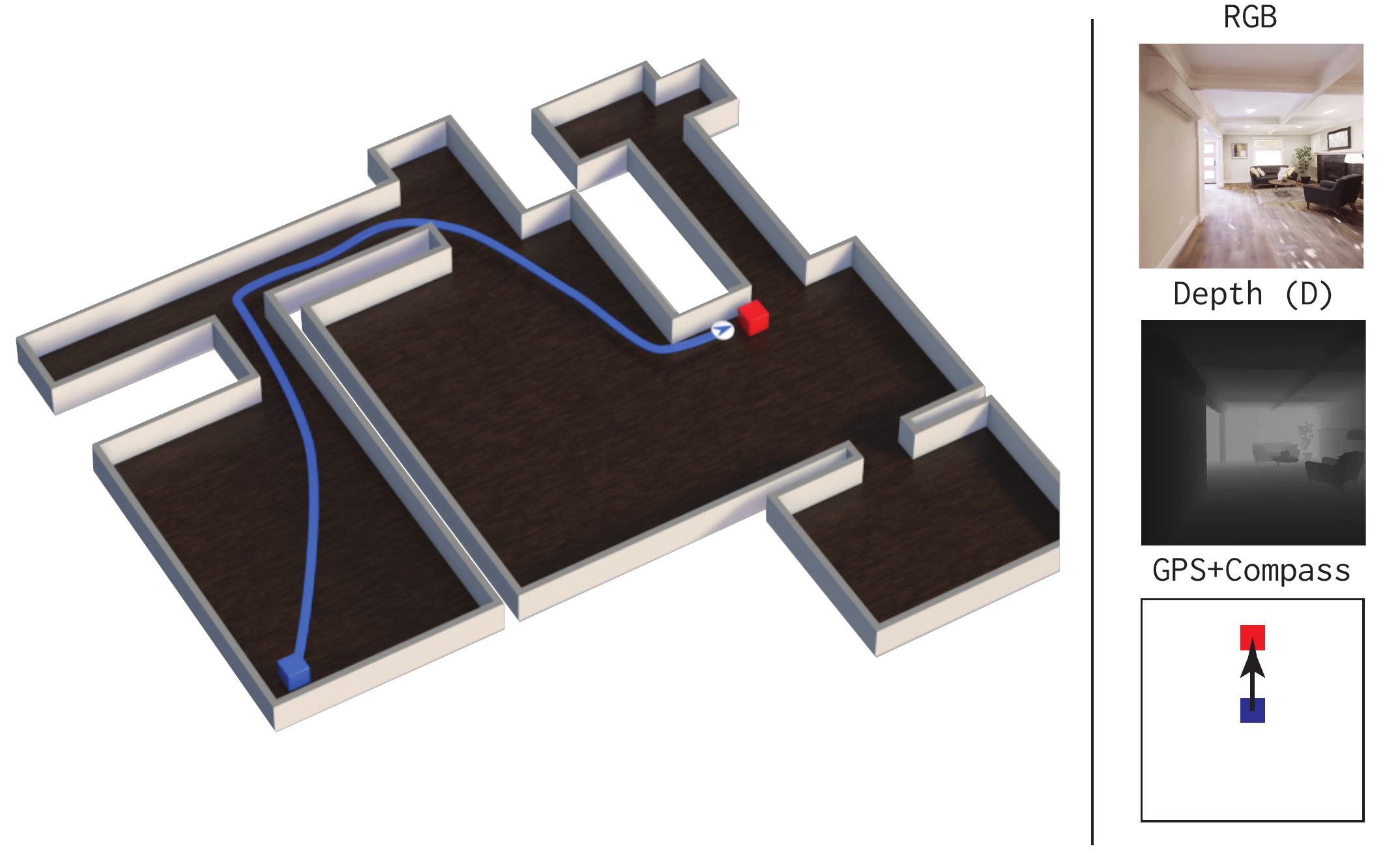}
    \caption{\xhdr{\pointnavfull}~\citep{anderson2018evaluation}. An agent is initialized in a novel environment (blue square) and task with navigation to a point specified relative to the start location (red square) -- \eg (5, 2) means go 5 meters forward and 2 meters right.  It must do so from egocentric inputs -- \rgbd and \gpscompass -- and without a map.}
    \label{fig:pointnav}
\end{figure}

In this rich space of tasks, \pointnavfull with \gpscompass has emerged as a test-bed problem due to its property of  `easy to get off the ground, but difficult to fully solve'. Specifically, \ddppo show that good performance can be obtained with under a week of GPU time but near-perfect performances currently requires \emph{half-a-year} of GPU time.
Moreover, findings and improvements in \pointnav with \gpscompass generally translate to other tasks (\pointnav without \gpscompass~\cite{ramakrishnan2020occupancy,datta2020integrating}, ObjectGoal navigation~\cite{objectnav,chaplot2020object}, RoomGoal navigation~\cite{narasimhan2020seeing}) and to navigation by real robots~\cite{habitatsim2real19arxiv}.

Given this interest, we present a systematic analysis of what matters and what what doesn't matter for learning \pointnav with \gpscompass on a limited sample or compute budget.  
We identify and discuss a number of ostensibly minor but significant design choices -- the advantage estimation procedure (a keep component in training), visual encoder architecture, and a hyper-parameter change -- that have large impacts on agent performance.

We examine these differences in two contexts, i) \emph{sample} efficiency, and ii) \emph{compute} efficiency.  To study sample efficiency, we train all agent variants for a fixed number of samples -- 75 million steps as represents a high but feasible number of samples~\cite{habitat19iccv,ye2020auxiliary,chaplot2020learning,sax2019learning,deitke2020robothor}.  To study compute efficiency, we ask a subtly different but important question:  \textbf{\emph{How far can we get with 1 GPU for 1 day?}}   Instead of training for a fixed number of samples, we train for a fixed amount of computation -- \ie comparisons at a different number of samples but the same amount of computation. 

Specifically, we contend that when that when training in simulation compute efficiency  should be an equally important objective. For instance, an agent architecture or training regime that \emph{increases} the number of samples required 2-fold but \emph{decreases} the compute required 6-fold would be desirable when training in simulation.  On the other hand, an architecture or training regime that \emph{reduces} the number of samples required 2-fold but \emph{increases} the compute required 6-fold would not be desirable.

Under these two objectives, we conduct an extensive set of experiments, cumulatively totalling over 50,000 GPU hours, and draw the following findings:

\begin{compactenum}[--]
\item \xhdr{Visual Encoder.} Relatively small CNNs are common place in visual navigation~\cite{ai2thor,gordon2019splitnet,sax2019learning,allenact,xia2018gibson}.  We examine using a moderately deep CNN (ResNet18~\citep{he2016resnet}) as \ddppo used a very deep to `solve' the task. We find that ResNet18 improves the sample efficiency of all agent variants and improve the compute efficiency of most -- it only harms the compute efficiency of  \depth-only agents.  ResNet18's improvements to sample efficiency outweigh its increased compute for \rgbd agents and considerably for \rgb agents.
\item \xhdr{Advantage Estimation.} Batch-wise normalizing advantage (a key component in policy optimization) is a common place `code-level' optimization~\cite{engstrom2020implementation} for advantage actor-critic methods however it is rarely discussed in the literature.  We present a systematic and controlled study of it in multi-environment goal-conditioned reinforcement learning. For the simple 3-layer CNN agents, we find it harms performance.  For the ResNet18 agents, we find it provides little to no benefit and it can make training unstable.    Given this, we recommend not batch-wise normalizing advantage.
\item \xhdr{Mini-Batch Size.}  We perform systematic experiments to examine the impact of learning mini-batch size and that large batches (\ie containing over 300 transitions) improves both the sample efficiency and compute efficiency for \emph{all} agent variants studied.
\end{compactenum}
These design choices to lead considerable and consistent improvements. Under a fixed sample budget, performance for \rgbd agents improves over the Habitat baselines by 8 SPL on Gibson (14\% relative improvement) and \emph{20} SPL on Matterport3D (38\% relative improvement). Under a fixed compute budget (1 GPU-day), performance for \rgbd agents improves by 19 SPL on Gibson (32\% relative improvement) and \emph{35} SPL on Matterport3D (220\% relative improvement). We hope our findings and recommendations will make serve to make the community's experiments more efficient.

\csection{Related Work}

\xhdr{Visual Navigation.} Training virtual robots for visual navigation has been a subject of much interest in recent years~\citep{embodiedqa,wijmans2019embodied,anderson2018vision,gordon2018iqa}. In this space, \pointnavfull has seen 
considerable interest
recently with works proposing to leverage hierarchical models~\citep{chaplot2020learning}, pre-trained representations~\citep{sax2019learning,gordon2019splitnet}, auxiliary-tasks~\cite{ye2020auxiliary}, or distribution and very deep networks~\citep{ddppo} to improve performance and/or sample efficiency.   %

\xhdr{Sample efficiency in model-free reinforcement learning (RL).}  Sample efficiency has been studied extensively in model-free RL, with works proposing new training methods~\citep{schulman2017ppo,haarnoja2018soft}, auxiliary losses~\citep{guo2018neural,guo2020bootstrap,srinivas2020curl}, intrinsic rewards~\citep{pathak2019self}, pre-training~\citep{sax2019learning}, and data-augmentation~\citep{kostrikov2020image,laskin_lee2020rad} to improve sample efficiency.  In comparison, we study how ostensibly minor changes impact sample efficiency and find they can have profound impacts.

\xhdr{Compute efficiency in model-free RL.} Compute efficiency has been studied in model-free RL in the context of distributed systems methods~\citep{ddppo,petrenko2020sample,espeholt2018impala,assran2019gossip}. We study compute efficiency with less compute (1 GPU for 1 day) and instead study changes to architecture or hyper-parameters.

\xhdr{Systematic studies of RL.} Closely related to our work is that of \citet{engstrom2020implementation}, who present systematic studies of how `code-level' optimizations impact perform for on-policy RL in Atari. Concurrently,  \citet{andrychowicz2020matters} present a systematic study of the impact of a wide variety of techniques on on-policy RL in DeepMind Control.  Of particular relevance to our work, they find mixed results batch-normalizing advantage.  We find this technique to often be harmful and their result helps to confirm that normalized advantage may not be necessary in general.  These works both perform their study in single-environment single-goal reinforcement learning problems (Atari and DeepMind Control).  We instead multi-environment (\ie agents trained in multiple houses) goal-conditioned (\ie agents trained to navigate to a provided goal) reinforcement learning, to the best of our knowledge this is the first such study.

\begin{figure*}
    \centering
    \includegraphics[width=0.95\textwidth]{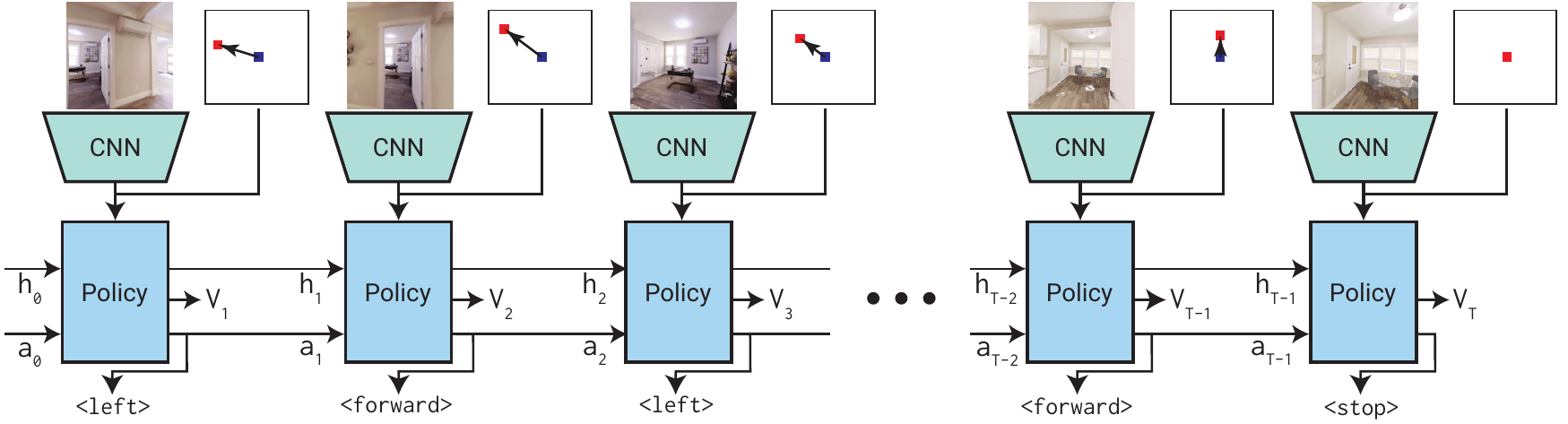}
    \caption{\xhdr{Agent Architecture.}  The architecture consists of two components, a visual encoder (parametrized by a CNN) and a policy (parametrized by an RNN).  At every timestep, the agent receives visual observations (\eg \rgb or \rgbd), use it's \gpscompass sensor to update the goal coordinates to be relative to its current location, and then predicts an action.  We compare the SimpleCNN agent from \citet{habitat19iccv} and the ResNet~\citep{he2016resnet} agent from \ddppo.}
    \label{fig:agent}
\end{figure*}

\csection{Task, Agent, and Training on a Sample\\Budget}
The starting point for our study is \savva and \ddppo, who both presented large advances in training agents for \pointnavfull -- the former showing that learned policies can outperform classical method and the latter showing that learned policies can `solve' this task when trained for 2.5 billion samples.
In this section we discuss their similarities and differences in task, agent, and training procedure.

\xhdr{\pointnavfull (\pointnav).}  \both train and evaluate agents for \pointnav~\citep{anderson2018evaluation} utilizing the AI Habitat platform~\citep{habitat19iccv}.  In \pointnav the agent is initialized at a random location in a previously unseen environment and then tasked with navigating to a point specified relative to the starting location -- \eg $(\delta x, \delta y)$ means go $\delta x$ meters forwards and $\delta y$ meters right. As shown in \reffig{fig:pointnav}, the agent is equipped with \rgb camera, a \depth sensor (or their combinations, \{\rgb, \depth, \texttt{RGB-D}\}) -- providing \texttt{256x256} egocentric color and/or depth observations at every time-step -- and a \gpscompass sensor -- providing position and orientation relative to the starting location at every time-step.

The agent has 4 primitive actions -- \texttt{move\_forward} (0.25m), \texttt{turn\_right}/\texttt{turn\_left} (10$^\circ$), and \texttt{stop}.
Performance is analyzed by two metrics -- 
\begin{inparaenum}[1)]
\item Success, whether or not the agent took \texttt{stop} within 0.2m of the goal, and 
\item Success weighted by (normalized inverse) Path Length (SPL)~\citep{anderson2018evaluation} -- a measure of success and efficiency.
\end{inparaenum}
Agents are trained and evaluated in either the (a) Matterport3D (MP3D) dataset~\citep{mp3d} -- comprised of 61/11 training/validation scenes --  or (b) \citet{habitat19iccv}'s curation of the Gibson dataset~\citep{xia2018gibson} -- containing 72/14 training/validation scenes.

\xhdr{Agent designs.}
\both employ similar agent architectures, shown in \reffig{fig:agent} -- the agent is divided into a visual encoder and policy, and utilizes its \gpscompass sensor to update the goal location to be relative to current location.  In  \citet{habitat19iccv}, the visual observation (\eg \rgb) is encoded by a 3-layer CNN (SimpleCNN) and  then concatenated with the goal (relative to current location) in polar coordinates ($[r, \theta]$).  The policy (a 1-layer GRU and a linear layer) then produces a softmax distribution over the action space and estimate of the value function.  

\ddppo instead use a large CNN (\eg ResNet50~\citep{he2016resnet}) to produce an embedding of the visual observation.  This is concatenated with a learnable embedding of the previous action taken and a learnable embedding of the goal.  Given the goal in polar coordinates $[r, \theta]$, the goal is first transformed to $[r, \cos(\theta), \sin(\theta)]$ and then projected to 32-d with a learned matrix to 1) handle the discontinuity at 180 degrees and 2) allow the agent to learn whatever normalization is sensible.  The policy is a 2-layer LSTM followed by a linear layer.
We change the 2-layer LSTM to a 1-layer GRU (matching the SimpleCNN agent) and use ResNet18 as the backbone visual encoder to reduce training time.  Further, plots in \ddppo show that the deeper CNNs are improve final performance but do not influence initial performance. We follow the same procedure as \ddppo to adapt ResNet18 -- reduce the number of output channels by half, add an initial \texttt{2x2-AvgPool}, replace BatchNorm with GroupNorm~\citep{wu2018group}, and replace the final global average pool with \texttt{3x3-Conv}+\texttt{Flatten} to produce a 2048-d vector.  As in \ddppo, for the \rgb and \rgbd agents, we normalize the visual inputs (subtract/divide by running-across-episodes mean/standard deviation).

\begin{table*}
    \setlength{\tabcolsep}{4pt}
    \centering
    \resizebox{\textwidth}{!}{
\begin{tabular}{l c c c cc c cc c cc c cc c}
\toprule
    & & & && \multicolumn{5}{c}{\textbf{SimpleCNN}} && \multicolumn{5}{c}{\textbf{ResNet18}} \\
    & & & && \multicolumn{2}{c}{\textbf{Gibson}} && \multicolumn{2}{c}{\textbf{Matterport3D}}  &&
             \multicolumn{2}{c}{\textbf{Gibson}} && \multicolumn{2}{c}{\textbf{Matterport3D}} \\
    \cmidrule{6-7} \cmidrule{9-10}
    \cmidrule{12-13} \cmidrule{15-16}
    \texttt{\#} & Sensors &  Norm Adv. & HyperParams &&
        \textbf{\texttt{Success}}~$\uparrow$ &  \textbf{\texttt{SPL}}~$\uparrow$ &&   \textbf{\texttt{Success}}~$\uparrow$ &  \textbf{\texttt{SPL}}~$\uparrow$ &&
        \textbf{\texttt{Success}}~$\uparrow$ &  \textbf{\texttt{SPL}}~$\uparrow$ &&   \textbf{\texttt{Success}}~$\uparrow$ &  \textbf{\texttt{SPL}}~$\uparrow$ \\

   \midrule 
    \texttt{1*} & \multirow{5}{*}{\textbf{\texttt{RGB}}} &  \checkmark & Set 1 && 72 & 54 && - & $\sim$28 && - & - && - & -\\ 
    \texttt{1} &  &  \checkmark & Set 1 && 69.2\scriptsize{$\pm$2.52} & 53.9\scriptsize{$\pm$1.39} && 34.8\scriptsize{$\pm$2.79} & 25.2\scriptsize{$\pm$1.76} && 71.6\scriptsize{$\pm$1.92} & 60.1\scriptsize{$\pm$1.99} && 46.5\scriptsize{$\pm$0.78} & 33.8\scriptsize{$\pm$1.23}\\ 
    \texttt{2} &  &  - & Set 1 && 75.3\scriptsize{$\pm$1.67} & 57.5\scriptsize{$\pm$2.00} && 34.2\scriptsize{$\pm$2.81} & 25.4\scriptsize{$\pm$2.44} && 79.6\scriptsize{$\pm$1.32} & 65.4\scriptsize{$\pm$1.10} && 46.8\scriptsize{$\pm$3.21} & 35.7\scriptsize{$\pm$0.92}\\ 
    \texttt{3} &  &  \checkmark & Set 2 && 14.2\scriptsize{$\pm$24.97} & 11.2\scriptsize{$\pm$19.56} && 00.0\scriptsize{$\pm$0.00} & 00.0\scriptsize{$\pm$0.00} && 79.3\scriptsize{$\pm$1.57} & 68.0\scriptsize{$\pm$1.26} && 53.1\scriptsize{$\pm$2.03} & 40.1\scriptsize{$\pm$0.86}\\ 
    \texttt{4} &  &  - & Set 2 && 78.5\scriptsize{$\pm$2.20} & 63.2\scriptsize{$\pm$1.76} && 55.1\scriptsize{$\pm$3.28} & 37.6\scriptsize{$\pm$1.77} && 81.8\scriptsize{$\pm$2.14} & 68.4\scriptsize{$\pm$1.38} && 50.8\scriptsize{$\pm$1.46} & 37.9\scriptsize{$\pm$1.40}\\ 

   \midrule 
    \texttt{5*} & \multirow{5}{*}{\textbf{\texttt{RGB-D}}} &  \checkmark & Set 1 && 78 & 69 && - & $\sim$42 && - & - && - & -\\ 
    \texttt{5} &  &  \checkmark & Set 1 && 79.8\scriptsize{$\pm$0.67} & 70.1\scriptsize{$\pm$1.32} && 62.0\scriptsize{$\pm$2.80} & 46.8\scriptsize{$\pm$2.10} && 87.0\scriptsize{$\pm$1.15} & 78.7\scriptsize{$\pm$1.62} && 73.3\scriptsize{$\pm$1.03} & 58.1\scriptsize{$\pm$1.39}\\ 
    \texttt{6} &  &  - & Set 1 && 84.4\scriptsize{$\pm$0.75} & 73.5\scriptsize{$\pm$0.67} && 71.4\scriptsize{$\pm$0.94} & 54.8\scriptsize{$\pm$2.39} && 87.2\scriptsize{$\pm$0.73} & 77.3\scriptsize{$\pm$0.37} && 74.6\scriptsize{$\pm$2.23} & 60.6\scriptsize{$\pm$1.49}\\ 
    \texttt{7} &  &  \checkmark & Set 2 && 85.1\scriptsize{$\pm$1.29} & 77.4\scriptsize{$\pm$1.32} && 55.5\scriptsize{$\pm$31.39} & 45.9\scriptsize{$\pm$26.00} && 87.4\scriptsize{$\pm$0.75} & 80.5\scriptsize{$\pm$0.77} && 78.8\scriptsize{$\pm$1.02} & 66.7\scriptsize{$\pm$1.07}\\ 
    \texttt{8} &  &  - & Set 2 && 87.3\scriptsize{$\pm$1.39} & 78.0\scriptsize{$\pm$0.79} && 75.2\scriptsize{$\pm$2.56} & 60.9\scriptsize{$\pm$1.82} && 88.6\scriptsize{$\pm$0.80} & 80.3\scriptsize{$\pm$0.36} && 78.3\scriptsize{$\pm$0.98} & 64.7\scriptsize{$\pm$1.05}\\ 

   \midrule 
    \texttt{9*} & \multirow{5}{*}{\textbf{\texttt{Depth}}} &  \checkmark & Set 1 && 86 & 78 && - & $\sim$55 && - & - && - & -\\ 
    \texttt{9} &  &  \checkmark & Set 1 && 85.2\scriptsize{$\pm$1.99} & 74.8\scriptsize{$\pm$3.23} && 70.5\scriptsize{$\pm$1.20} & 57.0\scriptsize{$\pm$0.79} && 89.9\scriptsize{$\pm$1.85} & 80.6\scriptsize{$\pm$1.98} && 76.0\scriptsize{$\pm$1.37} & 62.3\scriptsize{$\pm$1.44}\\ 
    \texttt{10} &  &  - & Set 1 && 87.4\scriptsize{$\pm$1.79} & 77.6\scriptsize{$\pm$1.35} && 75.7\scriptsize{$\pm$0.99} & 61.8\scriptsize{$\pm$1.16} && 91.3\scriptsize{$\pm$0.75} & 80.8\scriptsize{$\pm$0.49} && 77.5\scriptsize{$\pm$1.75} & 63.2\scriptsize{$\pm$1.01}\\ 
    \texttt{11} &  &  \checkmark & Set 2 && 88.6\scriptsize{$\pm$1.67} & 81.9\scriptsize{$\pm$0.93} && 78.3\scriptsize{$\pm$0.29} & 65.9\scriptsize{$\pm$0.50} && 91.8\scriptsize{$\pm$1.35} & 83.9\scriptsize{$\pm$0.58} && 81.8\scriptsize{$\pm$1.17} & 69.3\scriptsize{$\pm$1.49}\\ 
    \texttt{12} &  &  - & Set 2 && 93.1\scriptsize{$\pm$0.59} & 84.4\scriptsize{$\pm$0.44} && 80.1\scriptsize{$\pm$1.42} & 66.5\scriptsize{$\pm$1.14} && 93.0\scriptsize{$\pm$0.57} & 84.2\scriptsize{$\pm$0.55} && 80.5\scriptsize{$\pm$1.79} & 66.9\scriptsize{$\pm$0.97}\\ 
    \bottomrule 
\end{tabular}
}

	\caption{\xhdr{Results on \pointnav at 75 million.} Performance reported on the Gibson and Matterport3D validation sets (validation used to reduce exposure to test). Checkpoint selection done by validation SPL for each run independently. \texttt{*} denotes the results from \citet{habitat19iccv} for reference -- Matterport3D numbers taken from plots (and thus approximate), Gibson numbers from re-evaluation of released model weights.  We find that ResNet18 improves performance and normalized advantage can harm performance.  We also find that relatively minor changes in hyper-parameters results in large changes for SimpleCNN while ResNet18 without normalized advantage is more consistent. Mean and 95\% CI from 5 runs.}
	\label{tab:results}
\end{table*}

\xhdr{Advantage Estimation and Reward.}
One subtle difference between the training procedures 
is how advantage (a key component in PPO) is estimated.  
While  both \both  use GAE, \savva employ an additional trick known as normalized advantage --  the estimated advantages are normalized to have zero mean and unit variance per batch.  This is particularly important because this trick falls under the realm of `code-level optimizations'~\citep{engstrom2020implementation} and is not described in any papers using it, to the best of our knowledge.  While it is common place, effective, and (possibly) sensible single-environment single-goal settings, \ie Atari, it is unclear if this transfers to the multi-environment goal-conditioned setting of Habitat. \ddppo found this to introduce instabilities at their scale and did not use it.  We conduct systematic experiments to study its affect while controlling for others factors. 

\both use the same general reward structure.  Let $s_t$ be the agent's state at time $t$ and $a_t$ be the action taken, then reward is

\begin{equation}
    r(s_t, a_t) = \begin{cases}
    \beta \cdot \text{Success} & \text{if } a_t = \texttt{stop} \\
    -\Delta_{\text{GeoDist}}(s_t, a_t) - \lambda & \text{otherwise}
    \end{cases}
\end{equation}

\noindent
where $\Delta_{\text{GeoDist}}(s_t, a_t)$ is the change in geodesic distance to goal and $\lambda({=}0.01)$ is a slack penalty.\footnote{\ddppo weight the terminal reward by SPL.  We omit this for consistency and as it isn't influential at this scale of training.}  \citet{habitat19iccv} set $\beta{=}10.0$ and \ddppo set $\beta{=}2.5$ (no normalized advantage). 

\xhdr{Training.} \both use Proximal Policy Optimization (PPO)~\citep{schulman2017ppo} with Generalized Advantage Estimation (GAE)~\citep{schulman2016high}. Following \citet{habitat19iccv} we set utilize discount factor $\gamma{=}0.99$, GAE $\tau{=}0.95$, a rollout length of 128, and a learning rate of $2.5\times10^{-4}$.   We linearly decay the learning rate and PPO clip $\epsilon$ to 1/3rd of their initial values over the course of training.  For each PPO update, we collect 128$\times$6 steps of experience from 6 parallel instances of the simulator (all the same GPU and each with a different scene).  Recently, Habitat add support for texture compressed MP3D meshes.  These texture compressed meshes use 4$\times$-6$\times$ less GPU memory (\eg a reduction from 4GB to 700 MB)   With these, we are able to train agents on a \emph{single} 12GB Titan XP compared to the 4-GPU setup used in \citet{habitat19iccv} where 3 GPUs where used for rendering.

We consider two different settings for the PPO clip $\epsilon$ and the number of mini-batches per PPO epoch.
\begin{center}
    \begin{tabular}{c c c}
   
   & Set 1 & Set 2 \\
   \toprule
   PPO Clip ($\epsilon$) & 0.1 & 0.2 \\
   Mini-batches per epoch & 6\footnote{This was reported as 4 in \citet{habitat19iccv}, however, we confirmed with the authors and the implementation results in 6 for MP3D and 4 for Gibson.  We use their MP3D hyper-parameters on both MP3D and Gibson.}  & 2 \\
   \bottomrule
\end{tabular}
\end{center}
Set 1 replicates the Matterport3D hyper-parameters from \citet{habitat19iccv} exactly.  Set 2 results in hyper-parameters that are nearly identical to \ddppo. The only difference is 4 PPO epochs instead of 2 as \citet{habitat19iccv} used 4. We use the same number of PPO epochs for both hyper-parameter sets to isolate the effect of changing how many times a given step of experience is used to update the model.\footnote{We do not ablate 2 vs.~4 epochs to reduce the size of the grid.  The current grid required 240 agents and 40,000 GPU hours.}

We train agents under either a sample budget or compute budget. Under a sample budget, we match \citet{habitat19iccv} and use 75 million frames of experience.  Under a compute budget, we use 1-GPU day (discussed further in \refsec{sec:compute-training-protocol})

\xhdr{Evaluation Protocol.} Agents are evaluated on held out scenes from the validation set.  We do not use the test set to avoid over-fitting to the test-set.  During evaluation, the agent must navigate solely from its egocentric sensor (\ie \rgbd and \gpscompass) and reward is not given.

\begin{figure*}
    \centering
    \includegraphics[width=0.975\textwidth]{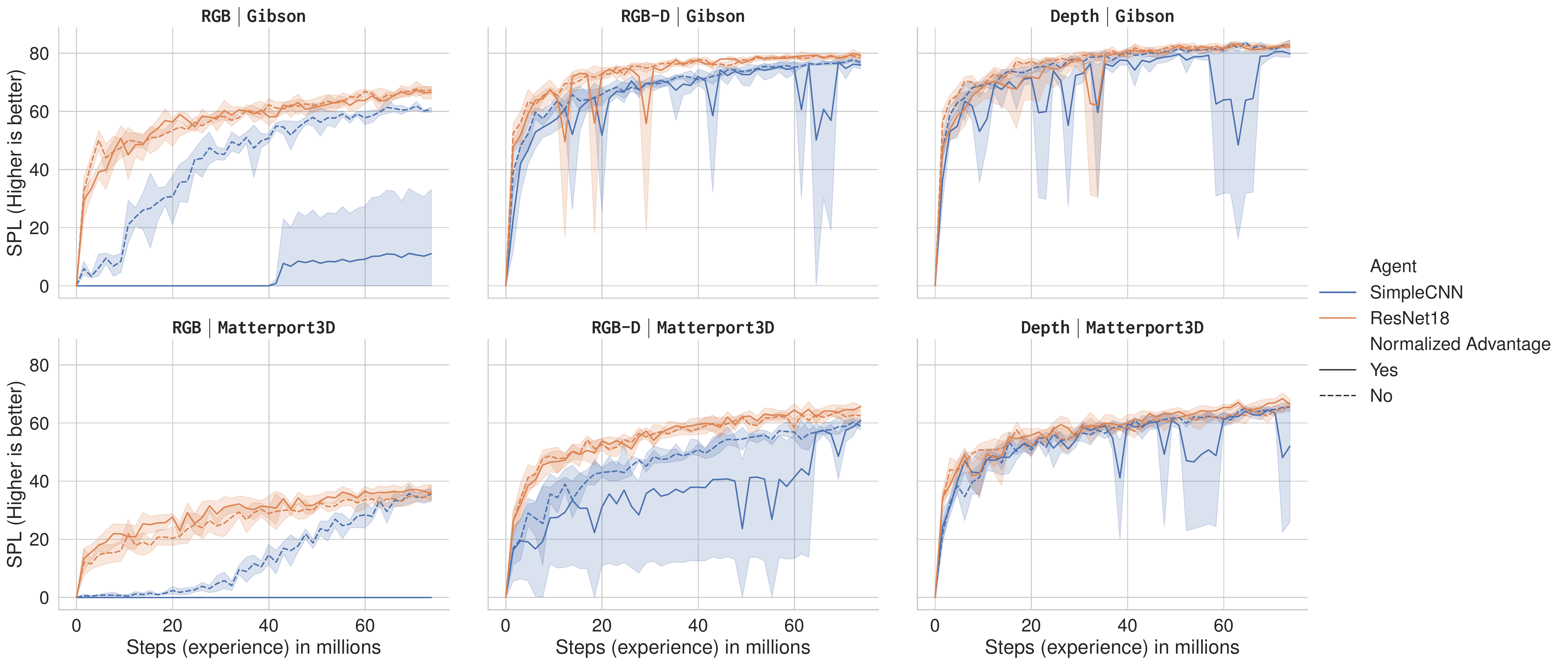}
    \caption{\xhdr{Performance (SPL; higher is better)} using  hyper-parameter set 2 as a function of experience.  Note the spurious dips in performance for the variants with normalize advantage.  Shading represents 95\% CI drawn over 5 runs.}
    \label{fig:spl-vs-steps}
\end{figure*}

\csection{Results on a Sample Budget}
\label{sec:results-75}

In this section we discuss our results when considering \emph{sample} efficiency.    \reftab{tab:results} shows our results.  The rows show the 12 different agent settings studied -- \{\rgb, \rgbd, \depth\}$\times$\{Hyper-Parameter Set 1, Hyper-Parameter Set 2\}$\times$\{normalized advantage, unnormalized advantage\}.  The columns show the 4 different settings each  agent is trained under -- \{Gibson, Matterport3D\}$\times$\{SimpleCNN, ResNet18\}.  We focus our analysis on hyper-parameter set 1 initially and then discuss hyper-parameter set 2.  We refer to changes in performance as \{+,-\}X/\{+,-\}Y SPL to indicate  an \{increase, decrease\} of X SPL on Gibson and a \{increase, decrease\} of Y SPL on Matterport3D.

\xhdr{Normalized advantage harms performance} (for SimpleCNN).  We find that normalized advantage harms performance for SimpleCNN in almost all cases and \emph{never} improves performance --  +6/+0 SPL for \rgb (row \texttt{2} vs.~\texttt{1}), +3/+8 SPL for \rgbd (\texttt{6} vs.~\texttt{5}), and +3/+5 SPL for \depth (\texttt{10} vs.~\texttt{9}).   For ResNet18, normalized advantage neither harms nor improves the performance of the best checkpoint by a statically significant margin. For both, normalized advantage introduces instability -- \ie agent performance will spuriously collapse before recovering shortly  (see \rgbd$\mid$Gibson in \reffig{fig:spl-vs-steps}).  Despite its prevalence, we find clear evidence that this method is harmful.

\xhdr{ResNet18 improves performance.}  The largest visible difference between \savva and \ddppo is the choice of visual encoder.  \savva use a simple 3-layer CNN that has its origins in Atari experiments~\citep{mnih2013playing} and contains none of the features of modern CNNs  -- \eg no skip-connections~\citep{he2016resnet} nor normalization layers~\citep{ioffe2015batch}. Due to the visual complexity of Gibson and Matterport, a better CNN improves performance considerably -- +8/+10 SPL for  \rgb (row \texttt{1}),  +4/+5 SPL for \rgbd (row \texttt{6}), and  +3/+0 SPL for \depth (row \texttt{10}).  As we transition from \rgb to \rgbd to \depth, the improvements due to ResNet18 decrease, indicating that \depth is already a highly conducive  visual representation for \pointnav with \gpscompass.

\xhdr{The gap between \rgbd and \depth closes.}  One intriguing trend of \savva is the difference in performance between the \rgbd and \depth agents, particularly on Matterport3D (gap of 13 SPL, row \texttt{5*} vs.~\texttt{9*}, left).  The \rgbd agent could clearly do better if it simply ignored \rgb but instead fails to learn to do so.  We find that given a better visual encoder, this gap closes considerably, to 2 SPL (row \texttt{6} vs.~\texttt{10}, right).  Specifically, the \rgbd performance on Matterport3D with ResNet18 and no normalized advantage is 61 SPL, while the performance with \depth is 63 SPL.

\xhdr{Hyper-parameter set 2 improves performance further.} For ResNet18 agents,
the relatively minor change to hyper-parameter set 2 improves SPL by at least 3 in all cases.
Interesting, we find that
SimpleCNN  benefits more than ResNet18 from hyper-parameter set 2. For \depth, SimpleCNN nearly matches the performance of the ResNet18 agent, indicating that \depth-only is an powerful inductive basis for \pointnav with \gpscompass.

\csection{Training on a Compute Budget}
\label{sec:compute-training-protocol}

\begin{table*}[t]
    \setlength{\tabcolsep}{4pt}
    \centering
    \resizebox{\textwidth}{!}{
\begin{tabular}{l c c c cc c cc c cc c cc c}
\toprule
    & & & && \multicolumn{5}{c}{\textbf{SimpleCNN}} && \multicolumn{5}{c}{\textbf{ResNet18}} \\
    & & & && \multicolumn{2}{c}{\textbf{Gibson}} && \multicolumn{2}{c}{\textbf{Matterport3D}}  &&
             \multicolumn{2}{c}{\textbf{Gibson}} && \multicolumn{2}{c}{\textbf{Matterport3D}} \\
    \cmidrule{6-7} \cmidrule{9-10}
    \cmidrule{12-13} \cmidrule{15-16}
    \texttt{\#} & Sensors &  Norm Adv. & HyperParams &&
        \textbf{\texttt{Success}}~$\uparrow$ &  \textbf{\texttt{SPL}}~$\uparrow$ &&   \textbf{\texttt{Success}}~$\uparrow$ &  \textbf{\texttt{SPL}}~$\uparrow$ &&
        \textbf{\texttt{Success}}~$\uparrow$ &  \textbf{\texttt{SPL}}~$\uparrow$ &&   \textbf{\texttt{Success}}~$\uparrow$ &  \textbf{\texttt{SPL}}~$\uparrow$ \\

   \midrule 
    \texttt{1} & \multirow{4}{*}{\textbf{\texttt{RGB}}} &  \checkmark & Set 1 && 11.3\scriptsize{$\pm$5.25} & 08.6\scriptsize{$\pm$3.88} && 01.0\scriptsize{$\pm$0.18} & 00.8\scriptsize{$\pm$0.15} && 59.0\scriptsize{$\pm$3.46} & 41.8\scriptsize{$\pm$4.50} && 24.8\scriptsize{$\pm$1.35} & 20.3\scriptsize{$\pm$1.43}\\ 
    \texttt{2} &  &  - & Set 1 && 34.0\scriptsize{$\pm$12.15} & 25.2\scriptsize{$\pm$8.58} && 01.1\scriptsize{$\pm$0.24} & 01.0\scriptsize{$\pm$0.21} && 50.9\scriptsize{$\pm$4.14} & 37.6\scriptsize{$\pm$2.37} && 15.3\scriptsize{$\pm$1.75} & 13.1\scriptsize{$\pm$1.56}\\ 
    \texttt{3} &  &  \checkmark & Set 2 && 00.0\scriptsize{$\pm$0.00} & 00.0\scriptsize{$\pm$0.00} && 00.0\scriptsize{$\pm$0.00} & 00.0\scriptsize{$\pm$0.00} && 63.1\scriptsize{$\pm$4.48} & 49.8\scriptsize{$\pm$3.79} && 29.4\scriptsize{$\pm$3.73} & 24.3\scriptsize{$\pm$1.91}\\ 
    \texttt{4} &  &  - & Set 2 && 46.2\scriptsize{$\pm$11.66} & 32.8\scriptsize{$\pm$9.00} && 02.3\scriptsize{$\pm$0.86} & 01.9\scriptsize{$\pm$0.66} && 68.3\scriptsize{$\pm$2.95} & 51.5\scriptsize{$\pm$1.63} && 22.3\scriptsize{$\pm$1.49} & 19.4\scriptsize{$\pm$0.65}\\ 

   \midrule 
    \texttt{5} & \multirow{4}{*}{\textbf{\texttt{RGB-D}}} &  \checkmark & Set 1 && 68.2\scriptsize{$\pm$1.97} & 51.2\scriptsize{$\pm$3.86} && 21.1\scriptsize{$\pm$4.45} & 15.7\scriptsize{$\pm$2.31} && 73.4\scriptsize{$\pm$1.53} & 59.8\scriptsize{$\pm$1.31} && 52.1\scriptsize{$\pm$1.56} & 38.3\scriptsize{$\pm$0.67}\\ 
    \texttt{6} &  &  - & Set 1 && 64.0\scriptsize{$\pm$5.42} & 49.1\scriptsize{$\pm$3.52} && 21.0\scriptsize{$\pm$7.72} & 16.0\scriptsize{$\pm$5.50} && 72.9\scriptsize{$\pm$3.04} & 56.6\scriptsize{$\pm$0.81} && 48.4\scriptsize{$\pm$2.83} & 40.2\scriptsize{$\pm$1.26}\\ 
    \texttt{7} &  &  \checkmark & Set 2 && 75.8\scriptsize{$\pm$3.83} & 65.0\scriptsize{$\pm$4.03} && 44.6\scriptsize{$\pm$25.38} & 34.4\scriptsize{$\pm$19.47} && 81.2\scriptsize{$\pm$1.67} & 68.8\scriptsize{$\pm$1.51} && 58.1\scriptsize{$\pm$5.24} & 47.0\scriptsize{$\pm$3.38}\\ 
    \texttt{8} &  &  - & Set 2 && 78.8\scriptsize{$\pm$1.47} & 66.2\scriptsize{$\pm$1.42} && 57.7\scriptsize{$\pm$2.43} & 42.1\scriptsize{$\pm$1.61} && 80.8\scriptsize{$\pm$2.47} & 68.0\scriptsize{$\pm$1.60} && 61.2\scriptsize{$\pm$2.27} & 50.4\scriptsize{$\pm$1.99}\\ 

   \midrule 
    \texttt{9} & \multirow{4}{*}{\textbf{\texttt{Depth}}} &  \checkmark & Set 1 && 77.5\scriptsize{$\pm$0.68} & 64.5\scriptsize{$\pm$2.49} && 49.1\scriptsize{$\pm$6.28} & 38.7\scriptsize{$\pm$2.80} && 78.0\scriptsize{$\pm$3.80} & 64.7\scriptsize{$\pm$4.37} && 55.1\scriptsize{$\pm$1.00} & 42.9\scriptsize{$\pm$1.25}\\ 
    \texttt{10} &  &  - & Set 1 && 75.0\scriptsize{$\pm$1.96} & 62.4\scriptsize{$\pm$2.08} && 46.7\scriptsize{$\pm$5.18} & 37.6\scriptsize{$\pm$3.54} && 74.2\scriptsize{$\pm$2.69} & 60.5\scriptsize{$\pm$1.09} && 54.8\scriptsize{$\pm$2.69} & 44.9\scriptsize{$\pm$2.26}\\ 
    \texttt{11} &  &  \checkmark & Set 2 && 81.4\scriptsize{$\pm$1.61} & 71.7\scriptsize{$\pm$2.20} && 67.1\scriptsize{$\pm$4.08} & 52.7\scriptsize{$\pm$3.10} && 80.8\scriptsize{$\pm$3.17} & 68.1\scriptsize{$\pm$4.41} && 62.2\scriptsize{$\pm$2.34} & 50.0\scriptsize{$\pm$1.90}\\ 
    \texttt{12} &  &  - & Set 2 && 87.1\scriptsize{$\pm$1.44} & 74.2\scriptsize{$\pm$0.67} && 67.1\scriptsize{$\pm$2.83} & 53.5\scriptsize{$\pm$2.65} && 83.0\scriptsize{$\pm$1.23} & 71.4\scriptsize{$\pm$1.63} && 63.2\scriptsize{$\pm$0.63} & 52.4\scriptsize{$\pm$0.95}\\ 
    \bottomrule 
\end{tabular}
}
	
	\caption{\xhdr{Results on \pointnav at 1 GPU-day.} Performance reported on the Gibson and Matterport3D validation sets. Under a fixed compute regime, we find that ResNet18 improves performance for \rgb and \rgbd agents while the increased number of samples achievable with SimpleCNN is beneficial for \depth agents. Mean and 95\% CI from 5 runs.}
	\label{tab:results-compute}
\end{table*}

We re-consider the comparisons above under a new context: we limit the amount of compute, not the number of samples.  We content that when training in simulation compute efficiency should be considered an equal counterpart.   In \refapx{apx:speed-improvements} we detail improvements we make to Habitat's compute utilization.

\begin{figure}
  \setlength{\tabcolsep}{4pt}
    \centering
    \resizebox{0.95\linewidth}{!}{
    \begin{tabular}{l l c c c}
    \toprule
    \shortstack[c]{Visual\\Encoder} & \shortstack[c]{Hyper\\parameters} &  \shortstack[c]{Throughput\\ (Samples per Second)}  & \shortstack[c]{Total\\(Millions)} & Factor \\
    \midrule
    ResNet18 & 1 & 100 &  8.0 & 1.00  \\
    ResNet18 & 2 & 120 &  10.0 & 1.25  \\
    SimpleCNN & 1 & 170 &  14.0 & 1.75  \\
    SimpleCNN & 2 & 200  & 16.0 & 2.00 \\
    \bottomrule
    \end{tabular}
   }
    \caption{\xhdr{Training speed} of the various settings considered this work. All numbers reported for \rgb agents on the Matterport3D~\citep{mp3d} dataset.  Times measure on a single 2080Ti GPU.}
    \label{tab:perf}
\end{figure}
\begin{table}

  \setlength{\tabcolsep}{4pt}
    \centering
    \begin{tabular}{l c c cc c c}
\toprule
& & && \multicolumn{2}{c}{\textbf{Gibson}} 
\\
\cmidrule{5-6} 
\texttt{\#} & Sensors &  Depth &&
    \textbf{\texttt{Success}}~$\uparrow$ &  \textbf{\texttt{SPL}}~$\uparrow$ %
    \\

   \midrule 
    \texttt{1} & \multirow{2}{*}{\textbf{\texttt{RGB}}} &  1  && 69.5\scriptsize{$\pm$1.76} & 53.3\scriptsize{$\pm$1.48} 
    \\ 
    \texttt{2} &  &  2  && 68.1\scriptsize{$\pm$2.29} & 50.2\scriptsize{$\pm$1.33} 
    \\ 

   \midrule 
    \texttt{3} & \multirow{2}{*}{\textbf{\texttt{RGB-D}}} &  1  && 78.8\scriptsize{$\pm$0.67} & 68.3\scriptsize{$\pm$0.37} 
    \\ 
    \texttt{4} &  &  2  && 79.3\scriptsize{$\pm$0.72} & 67.8\scriptsize{$\pm$1.14} 
    \\ 

    \bottomrule 
\end{tabular}

    \caption{\xhdr{Comparison of RNN Depth.} We find that a second RNN layer does not help \rgbd agents and may harm \rgb agents.  Mean and 95\% CI from 5 runs.}
    \label{tab:rnn-2-layer}
\end{table}

\xhdr{Samples Used.}
To enable clean comparisons with future work, we convert compute back into a number of samples, but encourage researches to stick the spirit of 1 GPU-day and adjust the number of samples if their method considerably changes the number of samples that can be obtained under this compute budget. In ~\reftab{tab:perf} we list the training throughput (in number of samples per second) for the agent configurations (both architecture and hyper-parameters) studied in this work.  We will compare agents trained for a different number of samples but all under a similar amount of compute.

\csection{Results on a Compute Budget}

We compare and contrast the trends found when training for a fixed number of samples in \refsec{sec:results-75} with a fixed compute budget of 1 GPU-day.  We examine the same variants as in \refsec{sec:results-75} and use the same method to describe results. \reftab{tab:results-compute} show these results.  We describe the trends bellow:

\xhdr{Hyper-parameter set 2} continues to outperform hyper-parameter set 1 for all variations.  This effect is now compounded by the $\sim$25\% more samples that can be collected under hyper-parameter set 2.  Under a fixed number of samples (10M for ResNet18 and 16M for SimleCNN), the average improvement when switching from hyper-parameter set 1 to 2 is 8.0{\scriptsize$\pm$2.75} SPL, which under a fixed compute budget (1 GPU-day), the average improvement is  9.3{\scriptsize$\pm$2.75} SPL.

\xhdr{Normalized advantage is inconsistent}, sometimes harming performance (-3.3/-2.4 SPL, \reftab{tab:results-compute}, row \texttt{11} vs.~\texttt{12}, ResNet18) and sometimes improving performance (-1.7/+4.9 SPL, \reftab{tab:results-compute}, row \texttt{3} vs.~\texttt{4}, ResNet18).  We hypothesis that early in training, normalized advantage  increases the weight of the policy loss, causing the agent to more optimize for initial performance at the cost of final performance.

\begin{table*}
  \setlength{\tabcolsep}{4pt}
    \centering
    
\resizebox{0.95\textwidth}{!}{
    \begin{tabular}{c ccc ccc ccc ccc ccc}
        \toprule
        & \multicolumn{14}{c}{\texttt{RolloutLength}} \\[4pt]
        &  \multicolumn{2}{c}{32} &&
            \multicolumn{2}{c}{48} &&
            \multicolumn{2}{c}{64} &&
            \multicolumn{2}{c}{96} &&
            \multicolumn{2}{c}{128} \\
            \cmidrule{2-3}
            \cmidrule{5-6}
            \cmidrule{8-9}
            \cmidrule{11-12}
            \cmidrule{14-15}
        \texttt{NumSim} &
            \textbf{\texttt{Success}}~$\uparrow$ &  \textbf{\texttt{SPL}}~$\uparrow$ &&   \textbf{\texttt{Success}}~$\uparrow$ &  \textbf{\texttt{SPL}}~$\uparrow$ &&
        \textbf{\texttt{Success}}~$\uparrow$ &  \textbf{\texttt{SPL}}~$\uparrow$ &&   \textbf{\texttt{Success}}~$\uparrow$ &  \textbf{\texttt{SPL}}~$\uparrow$ &&
        \textbf{\texttt{Success}}~$\uparrow$ &  \textbf{\texttt{SPL}}~$\uparrow$ \\
    \midrule
2 &08.4\scriptsize{$\pm$5.56} & 07.3\scriptsize{$\pm$4.78} && 12.2\scriptsize{$\pm$3.22} & 10.2\scriptsize{$\pm$2.52} && 18.2\scriptsize{$\pm$2.17} & 16.3\scriptsize{$\pm$1.97} && 23.0\scriptsize{$\pm$4.38} & 20.6\scriptsize{$\pm$4.15} && 28.4\scriptsize{$\pm$1.89} & 24.3\scriptsize{$\pm$1.42}\\ 
4 &21.9\scriptsize{$\pm$1.40} & 19.3\scriptsize{$\pm$1.04} && 22.3\scriptsize{$\pm$1.49} & 19.9\scriptsize{$\pm$1.43} && 26.1\scriptsize{$\pm$3.60} & 23.4\scriptsize{$\pm$2.96} && 29.6\scriptsize{$\pm$1.24} & 25.5\scriptsize{$\pm$1.09} && 34.4\scriptsize{$\pm$4.02} & 28.5\scriptsize{$\pm$4.27}\\ 
6 &23.9\scriptsize{$\pm$6.11} & 21.4\scriptsize{$\pm$5.34} && 26.2\scriptsize{$\pm$1.51} & 22.9\scriptsize{$\pm$1.16} && 32.1\scriptsize{$\pm$1.24} & 27.5\scriptsize{$\pm$1.16} && 33.5\scriptsize{$\pm$1.80} & 27.9\scriptsize{$\pm$0.94} && 38.5\scriptsize{$\pm$2.29} & 30.6\scriptsize{$\pm$1.37}\\ 
    \bottomrule 
\end{tabular}
}
    \caption{\xhdr{Hyper-parameter analysis.} Results of hyperpameter analysis at 40 million frames.   We find that, in general, performance improves as the number of steps collected per rollout (\texttt{NumSim}$\times$\texttt{RolloutLength}) increases and there isn't a significant trade-off between reducing either to reduce GPU memory. Mean and 95\% CI from 5 runs.}
    \label{tab:hparam-sweep}
\end{table*}

\xhdr{\rgb favors ResNet18, \depth-only favors \mbox{SimpleCNN}.} The largest difference due to the fixed compute regime is ResNet18 vs.~SimpleCNN.  Under the fixed sample regime, ResNet18 is a better choice regardless of visual input.  Under a fixed compute regime however, the significant increase in the number of samples that can be collected with SimpleCNN allows it to outperform ResNet18 for \depth-only agents in all cases except Matterport3D with hyper-parameter set 1. This is perhaps why researchers may have been using SimpleCNN~\cite{habitat19iccv,gordon2019splitnet,sax2019learning,allenact} -- under a given compute budget, it does perform better when the visual representation is already close to correct for the task.

For \rgb agents, ResNet18 continues to be a better choice -- so much so that a ResNet18 \rgb agent trained with hyper-parameter set 1 out-performs a SimpleCNN \rgb agent trained under hyper-parameter set 2, +3.8/+11.2 SPL (\reftab{tab:results-compute}, row \texttt{4} vs.~\texttt{2}), despite the \emph{2x} sample difference between the two.  Interestingly, for \rgbd agents, ResNet18 continues to outperform SimpleCNN, hinting that the primary benefit ResNet18 brings for \rgbd agents is the ability to more effectively learn to ignore the \rgb component.

\xhdr{Does RNN Depth Matter?}  In the context of a compute budget, we examine whether RNN depth matters.  Interestingly, RNN depth does not have a measure impact on the number of samples that can be obtained and thereby provides `free' additional network capacity.  We examine if this improves performance for \rgb and \rgbd agents on the Gibson dataset using ResNet18 and hyper-parameter set 2 (our most best setting for these agents).  \reftab{tab:rnn-2-layer} shows these results.  Interestingly, we find that the extra layer makes no difference for \rgbd and likely harms performance for \rgb.

\csection{Analysis}

In this section we provide additional analysis on two trends highlighted in our results, the impact of batch size and the negative impact of normalized advantage.

\csubsection{Batch Size}

One surprising trend from our analysis is that of hyper-parameter set 2 improves both sample efficiency \emph{and} compute efficiency.  The improvements in compute efficiency are intuitive, they are due to using a large mini-batch size during learning (which leads to better GPU utilization and less over-all gradient updates).  However, the improvements to sample efficiency are less intuitive.  To investigate, we systematically analyze different mini-batch sizes\footnote{Note that this experiment was performed after we had the results presented thus far and was not used to inform the choices of the hyper-parameter sets.}.

The number of steps of experience per mini-batch is defined as follows: Let \texttt{NumSim} be the number of parallel simulators, \texttt{RolloutLength} be the number of steps of experience collected from each per rollout, and \texttt{NumMiniBatch} be the number of PPO mini-batches, then the number of steps of experience per mini-batch  is $\frac{\texttt{NumSim}}{\texttt{NumMiniBatch}} \times \texttt{RolloutLength}$.
We fix \texttt{NumMiniBatch}{=}2  and examine all combinations of \texttt{NumSim} $\in\{2, 4, 6\}$ and \texttt{RolloutLength} $\in\{32, 48, 64, 96, 128\}$.  We examine their effect on \rgb ResNet18 agents without normalized advantage on the Matterport3D dataset as it is the most taxing sensor/dataset combination (lowest performance).  To examine sample efficiency, we train for 40 million steps of experience.  We note that the settings with shorter \texttt{RolloutLength} or lower \texttt{NumSim} are less compute efficient\footnote{The \texttt{NumSim}=2 \texttt{RolloutLength}=32 setting took 2 weeks!}.

We find that, overall, a larger batch size is more sample efficient (see the monontic improvement in performance in \reftab{tab:hparam-sweep}). In the context of \pointnav with 1-GPU, \emph{a larger batch size is  more sample efficient} despite reducing the total number of parameter updates and increasing the amount of experience gathered between updates.  This is likely due to reducing the variance~\cite{smith2017don} and increasing the number of positive examples in the mini-batch.

We find that \texttt{NumSim}=2 tends to under perform, for instance \texttt{NumSim}=2 \texttt{RolloutLength}=64 collects the same amount of experience per update as \texttt{NumSim}=4 \texttt{RolloutLength}=32 but results in -6 Success, -5 SPL, \reftab{tab:hparam-sweep}.  We hypothesis that \texttt{NumSim}=2 underperforms as, with \texttt{NumMiniBatch}{=}2, each batch has experience from the same environment, leading to highly correlated data within a batch. 
For the other values of \texttt{NumSim}, there isn't any particular trade-off between reducing \texttt{NumSim} or \texttt{RolloutLength}.

\csubsection{Normalized Advantage}

Despite its near ubiquity in  advantage actor-critic methods (like PPO), we find normalized advantage to be harmful.  To investigate this trend, we tested two possible hypothesis for why normalized advantage is harmful:
\begin{compactenum}[1)]
\item Division by a small standard deviation.  Early in training, advantages can be quite small and division by a small standard deviation would cause gradient spikes.
\item Variance in the estimate of normalization parameters later in training.  Later in training, advantages become larger in size (and thus larger variance), however, the normalization parameters are estimated per mini-batch and poorly estimated normalization parameters may lead to divergence.
\end{compactenum}
We test both these possibilities by 1) clipping the minimum value for the estimated standard deviation to 1, \ie $\sigma = \max(\sigma, 1)$ and 2) utilizing an exponential moving average of mean and variance.  This procedure is the same as that used in Mnih \etal~\cite{mnih2014neural}.

We examine this for \rgb agents on Gibson using Hyper-parameter Set 2 for both SimpleCNN and ResNet18 agents as it was harmful to the former and non-influential to the latter, results in \reftab{tab:norm-advant-variant}.  For ResNet18, this variant is not statistically different from per mini-batch or no normalization in either regime (52.6/66.4 SPL vs.~49.8/68.0 SPL vs.~51.5/68.4 SPL -- \reftab{tab:norm-advant-variant} row \texttt{2} vs.~\reftab{tab:results} row \texttt{3/4} vs.~\reftab{tab:results-compute} row \texttt{3}).  For SimpleCNN, this variant outperforms per mini-batch normalization in both regimes (40.1/61.8 SPL vs.~0.0/11.2). It outperforms no normalization in the compute-limited regime while both are on-par in the sample limited regime (32.8/63.2).  This indicates that a small variance likely explains the initial harm in performance.  However the instability later in training remains, indicating that neither effect fully explain this phenomena. This opens an interesting avenue for future work to develop an advantage normalization scheme that improves initial performance without causing instability.

\begin{table}

  \setlength{\tabcolsep}{4pt}
    \centering
  \resizebox{0.95\linewidth}{!}{
\begin{tabular}{l c c cc c}
\toprule
& & && \multicolumn{2}{c}{\textbf{Gibson}}  \\
\cmidrule{5-6} 
\texttt{\#} & Sensors &  Policy &&
    \textbf{\texttt{SPL}}@Compute~$\uparrow$ &  \textbf{\texttt{SPL}}@Sample~$\uparrow$ \\
   \midrule 
    \texttt{1} & \multirow{2}{*}{\textbf{\texttt{RGB}}} &  
    SimpleCNN  && 40.1\scriptsize{$\pm$3.31} & 61.8\scriptsize{$\pm$0.96} 
    \\ 
    \texttt{2} &  &  ResNet18  && 52.6{\scriptsize$\pm$2.08} & 66.4\scriptsize{$\pm$1.28} 
    \\ 
    \bottomrule 
\end{tabular}
}
    \caption{\xhdr{Second Normalized Advantage Method.} Results of a second variant of normalized advantage with hyper-parameter set 2.  While this method is not harmful performance, it still introduces instability.}
    \label{tab:norm-advant-variant}
\end{table}

\csection{Discussion}

In this work we have proposed that when training in simulation compute efficiency should be an equal counterpart to sample efficiency.  We argue that when training in simulation, the cost of a sample is due to the compute needed to generate it and learn from it, not something intrinsic to the sample itself.  When training in reality however, the cost to obtain a sample (monetary and time cost of the both the robot and its operator) dramatically outweighs the computation cost of learning from the sample, thus the focus has duly been put on sample efficiency (or when training in simulation as a direct stand-in for reality).  However, training in simulation, be it for evaluation in solely simulation, pre-training before fine-tuning in reality, or for zero-shot sim2real, is becoming increasingly popular and we argue that compute efficiency should be considered equally important in this context. 

We note that reproducibility of results under a  compute budget is challenging -- different simulators run at different speeds, existing simulators, GPU hardware, CUDA, and cuDNN get faster, \etc.  Given these challenges, comparisons under precisely the same compute budget is not feasible.  Instead, we encourage researchers to examine the the compute efficiency of their proposed method and compare against prior work under  \emph{similar} compute budgets.

We present experiments and analysis of how ostensibly minor changes influence both the sample efficiency and compute efficiency of embodied agents.
Boiling them down, we draw the following primary findings:

\xhdr{Use a deeper CNN, \eg ResNet18.}
Due to its long use in reinforcement learning for video games (i.e. Atari), the CNN from \citet{mnih2013playing} has also been adopted for visual navigation (it is the default in both Habitat~\citep{habitat19iccv} and AllenAct~\citep{allenact}). Our results show that this CNN is outperformed by ResNet18 -- not just in terms of accuracy (which is expected) but also in compute efficiency and sample efficiency, which is a surprising and counterintuitive finding -- priori we might expect larger models to be \emph{less} efficient, not more. This also goes against the trend in continuous control where larger CNNs harm performance without augmentation~\citep{kostrikov2020image}. 
We expect this trend will hold for tasks that rely even more on visual recognition, like ObjectGoal navigation~\citep{objectnav}, question answering~\citep{embodiedqa}, object rearrangement~\citep{rearrangement}, \etc.

\xhdr{Use large and diverse mini-batches.}
The effect of batch size for compute efficiency is perhaps expected, however, it is quite surprising for sample efficiency.  In the setting of learning in realistic simulators, the batch size is often constrained by GPU memory, limiting the feasible range one would search.  Our results show that even a seemingly large batch size of 3$\times$64 is not large enough.  
Second, we show that NumSim=2 consistently underperforms.  While this is a natural choice to reduce the GPU memory footprint of simulation, our results show that this is a suboptimal choice and other techniques need to be employed to fit within the GPU memory limit.  We expect these overall trends, larger batch sizes improve sample efficiency, and NumSim=2 underperforms, to hold true on other tasks as these control the gradient variance and diversity in the mini batch, which is not task specific.   Further, there is evidence that harder tasks require even larger batch sizes~\cite{mccandlish2018empirical}.

\xhdr{Do not use normalized advantage .}
Normalized advantage is a nearly ubiquitous trick in advantage actor critic methods that have been tuned for single environment reinforcement learning.  We show that this widely prevalent trick can be harmful for embodied navigation while providing little-to-no benefit when it isn't.  

We encourage researchers to compare against these agents trained with these tips and tricks and use them (where applicable) to ensure that advances build upon known best practices instead of rectifying for their absence.

\clearpage

\csection{Acknowledgements}
EW is supported in part by an ARCS fellowship.
The Georgia Tech effort was supported in part by NSF, AFRL, DARPA, ONR YIPs, ARO PECASE. The views and conclusions contained herein are those of the authors and should not be interpreted as necessarily representing the official policies or endorsements, either expressed or implied, of the U.S. Government, or any sponsor.

{\small
\bibliographystyle{cvpr/ieee_fullname}
\bibliography{bib/strings,bib/main}
}

\clearpage

\appendix
\renewcommand\thesection{\Alph{section}}
\setcounter{section}{0}
\renewcommand\thefigure{A\arabic{figure}}
\renewcommand\thetable{A\arabic{table}}
\setcounter{figure}{0}
\setcounter{table}{0}
\phantomsection

\iftoggle{supmain}{}{
\section{Additional Break Down}}

We provide additional break downs of our main results on sample efficiency at different sample budgets.

\section{Improvements to compute utilization.}  
\label{apx:speed-improvements}

Since our goal is maximize our performance given a fixed compute budget, we first make improvements to the habitat-baselines code-base to improve training throughput for \emph{all} models studied.\footnote{We will make code publicly available.} The largest improvement is due to utilizing a Double Buffered~\citep{petrenko2020sample}/Alternating-GPU~\citep{stooke2019rlpyt} sampler that allows for 20\% more samples to be collected under a given compute budget.  While collecting experience in the rollout, these sampling methods break the environments running in parallel into two groups.  Half the environments will simulate the result of taking the next action while the policy is executed to select the next action for the other half.  This allows network inference and simulation to be interleaved, leading to better usage of the PCIe bus, the fixed functions on the GPU that are dedicated to rendering, and a reduction in synchronization overhead. We make  additional minor improvements that result in another 10\%-15\% training speed.

\begin{table*}
    \setlength{\tabcolsep}{4pt}
    \centering
	\resizebox{\textwidth}{!}{
\begin{tabular}{l c c c cc c cc c cc c cc c}
\toprule
    & & & && \multicolumn{5}{c}{\textbf{SimpleCNN}} && \multicolumn{5}{c}{\textbf{ResNet18}} \\
    & & & && \multicolumn{2}{c}{\textbf{Gibson}} && \multicolumn{2}{c}{\textbf{Matterport3D}}  &&
             \multicolumn{2}{c}{\textbf{Gibson}} && \multicolumn{2}{c}{\textbf{Matterport3D}} \\
    \cmidrule{6-7} \cmidrule{9-10}
    \cmidrule{12-13} \cmidrule{15-16}
    \texttt{\#} & Sensors &  Norm Adv. & HyperParams &&
        \textbf{\texttt{Success}}~$\uparrow$ &  \textbf{\texttt{SPL}}~$\uparrow$ &&   \textbf{\texttt{Success}}~$\uparrow$ &  \textbf{\texttt{SPL}}~$\uparrow$ &&
        \textbf{\texttt{Success}}~$\uparrow$ &  \textbf{\texttt{SPL}}~$\uparrow$ &&   \textbf{\texttt{Success}}~$\uparrow$ &  \textbf{\texttt{SPL}}~$\uparrow$ \\

   \midrule 
    \texttt{1} & \multirow{4}{*}{\textbf{\texttt{RGB}}} &  \checkmark & Set 1 && 07.4\scriptsize{$\pm$0.93} & 05.9\scriptsize{$\pm$0.82} && 01.0\scriptsize{$\pm$0.18} & 00.8\scriptsize{$\pm$0.15} && 57.6\scriptsize{$\pm$3.81} & 42.3\scriptsize{$\pm$4.39} && 24.9\scriptsize{$\pm$1.34} & 20.5\scriptsize{$\pm$1.40}\\ 
    \texttt{2} &  &  - & Set 1 && 11.1\scriptsize{$\pm$2.39} & 08.7\scriptsize{$\pm$1.71} && 01.1\scriptsize{$\pm$0.24} & 00.9\scriptsize{$\pm$0.22} && 57.1\scriptsize{$\pm$1.69} & 41.4\scriptsize{$\pm$0.94} && 19.0\scriptsize{$\pm$5.31} & 15.6\scriptsize{$\pm$3.39}\\ 
    \texttt{3} &  &  \checkmark & Set 2 && 00.0\scriptsize{$\pm$0.00} & 00.0\scriptsize{$\pm$0.00} && 00.0\scriptsize{$\pm$0.00} & 00.0\scriptsize{$\pm$0.00} && 63.1\scriptsize{$\pm$4.48} & 49.8\scriptsize{$\pm$3.79} && 29.4\scriptsize{$\pm$3.73} & 24.3\scriptsize{$\pm$1.91}\\ 
    \texttt{4} &  &  - & Set 2 && 15.7\scriptsize{$\pm$2.07} & 11.6\scriptsize{$\pm$1.01} && 01.7\scriptsize{$\pm$0.88} & 01.4\scriptsize{$\pm$0.62} && 68.3\scriptsize{$\pm$2.95} & 51.5\scriptsize{$\pm$1.63} && 22.3\scriptsize{$\pm$1.49} & 19.4\scriptsize{$\pm$0.65}\\ 

   \midrule 
    \texttt{5} & \multirow{4}{*}{\textbf{\texttt{RGB-D}}} &  \checkmark & Set 1 && 60.4\scriptsize{$\pm$9.60} & 46.1\scriptsize{$\pm$6.61} && 07.6\scriptsize{$\pm$2.41} & 06.1\scriptsize{$\pm$2.00} && 74.3\scriptsize{$\pm$1.10} & 61.1\scriptsize{$\pm$0.82} && 51.9\scriptsize{$\pm$1.58} & 39.5\scriptsize{$\pm$0.97}\\ 
    \texttt{6} &  &  - & Set 1 && 48.6\scriptsize{$\pm$11.29} & 36.9\scriptsize{$\pm$9.19} && 07.1\scriptsize{$\pm$1.05} & 05.8\scriptsize{$\pm$0.90} && 72.1\scriptsize{$\pm$1.99} & 57.4\scriptsize{$\pm$1.64} && 49.0\scriptsize{$\pm$2.67} & 40.7\scriptsize{$\pm$1.04}\\ 
    \texttt{7} &  &  \checkmark & Set 2 && 69.1\scriptsize{$\pm$4.52} & 57.6\scriptsize{$\pm$4.35} && 44.5\scriptsize{$\pm$19.57} & 34.6\scriptsize{$\pm$15.24} && 81.2\scriptsize{$\pm$1.67} & 68.8\scriptsize{$\pm$1.51} && 58.1\scriptsize{$\pm$5.24} & 47.0\scriptsize{$\pm$3.38}\\ 
    \texttt{8} &  &  - & Set 2 && 76.8\scriptsize{$\pm$1.83} & 61.4\scriptsize{$\pm$0.87} && 52.5\scriptsize{$\pm$4.43} & 37.5\scriptsize{$\pm$3.03} && 80.8\scriptsize{$\pm$2.47} & 68.0\scriptsize{$\pm$1.60} && 61.2\scriptsize{$\pm$2.27} & 50.4\scriptsize{$\pm$1.99}\\ 

   \midrule 
    \texttt{9} & \multirow{4}{*}{\textbf{\texttt{Depth}}} &  \checkmark & Set 1 && 70.3\scriptsize{$\pm$2.46} & 57.3\scriptsize{$\pm$2.06} && 41.1\scriptsize{$\pm$7.52} & 32.8\scriptsize{$\pm$5.25} && 80.4\scriptsize{$\pm$3.20} & 67.7\scriptsize{$\pm$2.15} && 56.4\scriptsize{$\pm$0.97} & 43.2\scriptsize{$\pm$1.28}\\ 
    \texttt{10} &  &  - & Set 1 && 71.5\scriptsize{$\pm$1.51} & 57.1\scriptsize{$\pm$1.02} && 33.5\scriptsize{$\pm$5.24} & 26.7\scriptsize{$\pm$3.16} && 76.0\scriptsize{$\pm$1.91} & 62.2\scriptsize{$\pm$1.28} && 55.1\scriptsize{$\pm$2.26} & 45.3\scriptsize{$\pm$1.72}\\ 
    \texttt{11} &  &  \checkmark & Set 2 && 77.7\scriptsize{$\pm$2.03} & 67.1\scriptsize{$\pm$2.19} && 63.2\scriptsize{$\pm$4.25} & 49.8\scriptsize{$\pm$2.99} && 80.8\scriptsize{$\pm$3.17} & 68.1\scriptsize{$\pm$4.41} && 62.2\scriptsize{$\pm$2.34} & 50.0\scriptsize{$\pm$1.90}\\ 
    \texttt{12} &  &  - & Set 2 && 82.4\scriptsize{$\pm$2.93} & 70.1\scriptsize{$\pm$0.60} && 60.1\scriptsize{$\pm$3.69} & 44.2\scriptsize{$\pm$2.12} && 83.0\scriptsize{$\pm$1.23} & 71.4\scriptsize{$\pm$1.63} && 63.2\scriptsize{$\pm$0.63} & 52.4\scriptsize{$\pm$0.95}\\ 
    \bottomrule 
\end{tabular}
}
	\caption{\xhdr{Results on \pointnav at 10 million.} Same as \iftoggle{supmain}{Tab. 1}{\reftab{tab:results}} but at 10 million steps.}
	\label{tab:results-at-10}
\end{table*}

\begin{table*}
    \setlength{\tabcolsep}{4pt}
    \centering
	\resizebox{\textwidth}{!}{
\begin{tabular}{l c c c cc c cc c cc c cc c}
\toprule
    & & & && \multicolumn{5}{c}{\textbf{SimpleCNN}} && \multicolumn{5}{c}{\textbf{ResNet18}} \\
    & & & && \multicolumn{2}{c}{\textbf{Gibson}} && \multicolumn{2}{c}{\textbf{Matterport3D}}  &&
             \multicolumn{2}{c}{\textbf{Gibson}} && \multicolumn{2}{c}{\textbf{Matterport3D}} \\
    \cmidrule{6-7} \cmidrule{9-10}
    \cmidrule{12-13} \cmidrule{15-16}
    \texttt{\#} & Sensors &  Norm Adv. & HyperParams &&
        \textbf{\texttt{Success}}~$\uparrow$ &  \textbf{\texttt{SPL}}~$\uparrow$ &&   \textbf{\texttt{Success}}~$\uparrow$ &  \textbf{\texttt{SPL}}~$\uparrow$ &&
        \textbf{\texttt{Success}}~$\uparrow$ &  \textbf{\texttt{SPL}}~$\uparrow$ &&   \textbf{\texttt{Success}}~$\uparrow$ &  \textbf{\texttt{SPL}}~$\uparrow$ \\

   \midrule 
    \texttt{1} & \multirow{4}{*}{\textbf{\texttt{RGB}}} &  \checkmark & Set 1 && 31.7\scriptsize{$\pm$10.29} & 23.3\scriptsize{$\pm$7.46} && 01.7\scriptsize{$\pm$0.98} & 01.4\scriptsize{$\pm$0.75} && 63.1\scriptsize{$\pm$3.31} & 47.0\scriptsize{$\pm$1.85} && 32.3\scriptsize{$\pm$2.95} & 25.7\scriptsize{$\pm$1.27}\\ 
    \texttt{2} &  &  - & Set 1 && 47.4\scriptsize{$\pm$3.26} & 33.0\scriptsize{$\pm$2.71} && 01.4\scriptsize{$\pm$0.18} & 01.2\scriptsize{$\pm$0.14} && 65.1\scriptsize{$\pm$2.43} & 50.0\scriptsize{$\pm$1.27} && 25.2\scriptsize{$\pm$2.63} & 20.4\scriptsize{$\pm$1.47}\\ 
    \texttt{3} &  &  \checkmark & Set 2 && 00.0\scriptsize{$\pm$0.00} & 00.0\scriptsize{$\pm$0.00} && 00.0\scriptsize{$\pm$0.00} & 00.0\scriptsize{$\pm$0.00} && 73.3\scriptsize{$\pm$3.69} & 58.4\scriptsize{$\pm$2.67} && 36.2\scriptsize{$\pm$2.12} & 30.0\scriptsize{$\pm$1.30}\\ 
    \texttt{4} &  &  - & Set 2 && 52.9\scriptsize{$\pm$5.24} & 37.2\scriptsize{$\pm$5.13} && 03.3\scriptsize{$\pm$1.20} & 02.8\scriptsize{$\pm$0.98} && 70.2\scriptsize{$\pm$1.28} & 56.0\scriptsize{$\pm$0.97} && 29.4\scriptsize{$\pm$3.56} & 25.1\scriptsize{$\pm$2.35}\\ 

   \midrule 
    \texttt{5} & \multirow{4}{*}{\textbf{\texttt{RGB-D}}} &  \checkmark & Set 1 && 69.5\scriptsize{$\pm$1.30} & 56.2\scriptsize{$\pm$2.03} && 37.2\scriptsize{$\pm$8.69} & 28.2\scriptsize{$\pm$5.87} && 80.5\scriptsize{$\pm$1.58} & 70.4\scriptsize{$\pm$1.66} && 61.3\scriptsize{$\pm$2.39} & 48.1\scriptsize{$\pm$2.60}\\ 
    \texttt{6} &  &  - & Set 1 && 70.0\scriptsize{$\pm$2.65} & 55.0\scriptsize{$\pm$2.20} && 27.7\scriptsize{$\pm$7.42} & 21.7\scriptsize{$\pm$5.39} && 78.9\scriptsize{$\pm$2.79} & 67.0\scriptsize{$\pm$1.83} && 63.1\scriptsize{$\pm$2.83} & 50.1\scriptsize{$\pm$1.55}\\ 
    \texttt{7} &  &  \checkmark & Set 2 && 78.5\scriptsize{$\pm$3.71} & 66.9\scriptsize{$\pm$3.23} && 45.6\scriptsize{$\pm$25.96} & 34.4\scriptsize{$\pm$19.49} && 85.5\scriptsize{$\pm$0.91} & 75.2\scriptsize{$\pm$0.84} && 65.5\scriptsize{$\pm$2.60} & 54.0\scriptsize{$\pm$0.90}\\ 
    \texttt{8} &  &  - & Set 2 && 80.4\scriptsize{$\pm$0.91} & 67.2\scriptsize{$\pm$1.56} && 59.2\scriptsize{$\pm$3.33} & 45.0\scriptsize{$\pm$2.15} && 85.4\scriptsize{$\pm$1.20} & 73.9\scriptsize{$\pm$1.00} && 65.7\scriptsize{$\pm$2.02} & 55.1\scriptsize{$\pm$1.17}\\ 

   \midrule 
    \texttt{9} & \multirow{4}{*}{\textbf{\texttt{Depth}}} &  \checkmark & Set 1 && 76.7\scriptsize{$\pm$2.18} & 66.5\scriptsize{$\pm$1.78} && 55.3\scriptsize{$\pm$3.53} & 42.4\scriptsize{$\pm$2.53} && 83.2\scriptsize{$\pm$1.68} & 74.0\scriptsize{$\pm$1.26} && 60.7\scriptsize{$\pm$1.24} & 48.3\scriptsize{$\pm$1.54}\\ 
    \texttt{10} &  &  - & Set 1 && 77.0\scriptsize{$\pm$2.54} & 65.2\scriptsize{$\pm$1.92} && 59.1\scriptsize{$\pm$5.11} & 45.8\scriptsize{$\pm$3.94} && 82.5\scriptsize{$\pm$1.68} & 70.3\scriptsize{$\pm$0.76} && 64.7\scriptsize{$\pm$2.37} & 52.2\scriptsize{$\pm$1.68}\\ 
    \texttt{11} &  &  \checkmark & Set 2 && 82.9\scriptsize{$\pm$1.65} & 73.0\scriptsize{$\pm$1.94} && 68.2\scriptsize{$\pm$3.47} & 54.4\scriptsize{$\pm$1.92} && 83.4\scriptsize{$\pm$3.62} & 73.6\scriptsize{$\pm$6.26} && 70.7\scriptsize{$\pm$1.86} & 57.3\scriptsize{$\pm$1.50}\\ 
    \texttt{12} &  &  - & Set 2 && 85.8\scriptsize{$\pm$1.70} & 75.3\scriptsize{$\pm$0.39} && 69.0\scriptsize{$\pm$2.53} & 54.6\scriptsize{$\pm$2.09} && 87.8\scriptsize{$\pm$1.33} & 78.2\scriptsize{$\pm$1.02} && 70.0\scriptsize{$\pm$2.55} & 58.3\scriptsize{$\pm$1.61}\\ 
    \bottomrule 
\end{tabular}
}
	\caption{\xhdr{Results on \pointnav at 20 million.} Same as \iftoggle{supmain}{Tab. 1}{\reftab{tab:results}} but at 20 million steps.}
	\label{tab:results-at-20}
\end{table*}
\begin{table*}
    \setlength{\tabcolsep}{4pt}
    \centering
	\resizebox{\textwidth}{!}{
\begin{tabular}{l c c c cc c cc c cc c cc c}
\toprule
    & & & && \multicolumn{5}{c}{\textbf{SimpleCNN}} && \multicolumn{5}{c}{\textbf{ResNet18}} \\
    & & & && \multicolumn{2}{c}{\textbf{Gibson}} && \multicolumn{2}{c}{\textbf{Matterport3D}}  &&
             \multicolumn{2}{c}{\textbf{Gibson}} && \multicolumn{2}{c}{\textbf{Matterport3D}} \\
    \cmidrule{6-7} \cmidrule{9-10}
    \cmidrule{12-13} \cmidrule{15-16}
    \texttt{\#} & Sensors &  Norm Adv. & HyperParams &&
        \textbf{\texttt{Success}}~$\uparrow$ &  \textbf{\texttt{SPL}}~$\uparrow$ &&   \textbf{\texttt{Success}}~$\uparrow$ &  \textbf{\texttt{SPL}}~$\uparrow$ &&
        \textbf{\texttt{Success}}~$\uparrow$ &  \textbf{\texttt{SPL}}~$\uparrow$ &&   \textbf{\texttt{Success}}~$\uparrow$ &  \textbf{\texttt{SPL}}~$\uparrow$ \\

   \midrule 
    \texttt{1} & \multirow{4}{*}{\textbf{\texttt{RGB}}} &  \checkmark & Set 1 && 57.8\scriptsize{$\pm$6.23} & 43.8\scriptsize{$\pm$3.43} && 07.6\scriptsize{$\pm$3.00} & 06.2\scriptsize{$\pm$2.39} && 68.2\scriptsize{$\pm$2.87} & 52.7\scriptsize{$\pm$2.87} && 41.4\scriptsize{$\pm$1.61} & 30.1\scriptsize{$\pm$1.10}\\ 
    \texttt{2} &  &  - & Set 1 && 67.2\scriptsize{$\pm$1.66} & 49.0\scriptsize{$\pm$2.41} && 06.7\scriptsize{$\pm$2.58} & 05.5\scriptsize{$\pm$2.21} && 72.9\scriptsize{$\pm$2.27} & 55.5\scriptsize{$\pm$1.64} && 36.5\scriptsize{$\pm$2.79} & 28.1\scriptsize{$\pm$2.68}\\ 
    \texttt{3} &  &  \checkmark & Set 2 && 00.0\scriptsize{$\pm$0.01} & 00.0\scriptsize{$\pm$0.01} && 00.0\scriptsize{$\pm$0.00} & 00.0\scriptsize{$\pm$0.00} && 77.1\scriptsize{$\pm$1.76} & 63.0\scriptsize{$\pm$1.40} && 44.8\scriptsize{$\pm$5.85} & 35.3\scriptsize{$\pm$2.41}\\ 
    \texttt{4} &  &  - & Set 2 && 72.6\scriptsize{$\pm$1.32} & 53.6\scriptsize{$\pm$2.27} && 21.5\scriptsize{$\pm$5.32} & 15.4\scriptsize{$\pm$3.02} && 77.0\scriptsize{$\pm$1.11} & 62.9\scriptsize{$\pm$1.24} && 38.9\scriptsize{$\pm$1.53} & 32.6\scriptsize{$\pm$0.95}\\ 

   \midrule 
    \texttt{5} & \multirow{4}{*}{\textbf{\texttt{RGB-D}}} &  \checkmark & Set 1 && 76.6\scriptsize{$\pm$1.06} & 64.9\scriptsize{$\pm$0.56} && 53.3\scriptsize{$\pm$2.86} & 39.5\scriptsize{$\pm$1.81} && 84.2\scriptsize{$\pm$1.27} & 75.6\scriptsize{$\pm$1.08} && 66.2\scriptsize{$\pm$3.17} & 52.6\scriptsize{$\pm$1.28}\\ 
    \texttt{6} &  &  - & Set 1 && 80.0\scriptsize{$\pm$2.47} & 66.3\scriptsize{$\pm$2.23} && 56.2\scriptsize{$\pm$3.86} & 42.2\scriptsize{$\pm$1.65} && 84.9\scriptsize{$\pm$1.56} & 73.6\scriptsize{$\pm$1.13} && 67.6\scriptsize{$\pm$1.65} & 54.3\scriptsize{$\pm$0.98}\\ 
    \texttt{7} &  &  \checkmark & Set 2 && 81.4\scriptsize{$\pm$1.30} & 72.9\scriptsize{$\pm$1.87} && 50.9\scriptsize{$\pm$28.95} & 40.4\scriptsize{$\pm$23.02} && 86.4\scriptsize{$\pm$1.06} & 77.9\scriptsize{$\pm$0.55} && 72.8\scriptsize{$\pm$2.04} & 61.0\scriptsize{$\pm$1.32}\\ 
    \texttt{8} &  &  - & Set 2 && 84.4\scriptsize{$\pm$0.85} & 73.0\scriptsize{$\pm$1.07} && 66.6\scriptsize{$\pm$2.77} & 51.6\scriptsize{$\pm$2.00} && 86.5\scriptsize{$\pm$0.85} & 78.0\scriptsize{$\pm$0.42} && 72.2\scriptsize{$\pm$1.74} & 60.5\scriptsize{$\pm$0.77}\\ 

   \midrule 
    \texttt{9} & \multirow{4}{*}{\textbf{\texttt{Depth}}} &  \checkmark & Set 1 && 79.8\scriptsize{$\pm$1.41} & 71.0\scriptsize{$\pm$1.60} && 64.7\scriptsize{$\pm$1.92} & 51.1\scriptsize{$\pm$1.73} && 86.6\scriptsize{$\pm$1.02} & 77.6\scriptsize{$\pm$2.42} && 70.4\scriptsize{$\pm$2.99} & 57.2\scriptsize{$\pm$2.09}\\ 
    \texttt{10} &  &  - & Set 1 && 83.1\scriptsize{$\pm$2.08} & 72.8\scriptsize{$\pm$1.76} && 69.4\scriptsize{$\pm$4.29} & 53.7\scriptsize{$\pm$3.01} && 87.2\scriptsize{$\pm$1.19} & 76.4\scriptsize{$\pm$1.37} && 70.6\scriptsize{$\pm$1.22} & 57.6\scriptsize{$\pm$1.19}\\ 
    \texttt{11} &  &  \checkmark & Set 2 && 86.6\scriptsize{$\pm$0.70} & 78.8\scriptsize{$\pm$0.64} && 74.4\scriptsize{$\pm$0.92} & 61.6\scriptsize{$\pm$0.93} && 90.0\scriptsize{$\pm$1.42} & 81.4\scriptsize{$\pm$0.80} && 74.9\scriptsize{$\pm$1.13} & 62.7\scriptsize{$\pm$1.44}\\ 
    \texttt{12} &  &  - & Set 2 && 90.2\scriptsize{$\pm$1.63} & 80.6\scriptsize{$\pm$1.13} && 73.7\scriptsize{$\pm$2.66} & 60.0\scriptsize{$\pm$1.59} && 90.8\scriptsize{$\pm$0.96} & 81.1\scriptsize{$\pm$1.18} && 72.7\scriptsize{$\pm$1.76} & 62.1\scriptsize{$\pm$1.70}\\ 
    \bottomrule 
\end{tabular}
}
	\caption{\xhdr{Results on \pointnav at 40 million.} Same as \iftoggle{supmain}{Tab. 1}{\reftab{tab:results}} but at 40 million steps.}
	\label{tab:results-at-40}
\end{table*}

\begin{table*}
    \centering
    \begin{tabular}{l c}
    \toprule
    \multicolumn{2}{c}{PPO Parameters} \\
    \midrule
    PPO Epochs & 4 \\
    PPO Mini-Batches & 2 \\
    PPO Clip & 0.2 \\
    Advantage normalization & No \\
    $\gamma$ & 0.99 \\
    GAE-$\lambda$~\citep{schulman2016high} & 0.95 \\
    Learning rate &   $2.5\times 10^{-4}$ \\
    Max gradient norm & $0.5$ \\
    Optimizer & Adam~\citet{kingma2015adam} \\
    Number of Environemnts (\texttt{NumSim}) & 6 \\
    Rollout length (\texttt{RolloutLength}) & 128 \\
    \bottomrule
    \end{tabular}
    \caption{Base hyper-parameters (corresponds to hyper-parameter set 2 without normalized advantage).  All experiments use these parameters unless state otherwise.}
    \label{tab:hparams}
\end{table*}

\end{document}